\documentclass{llncs}

\usepackage{graphicx}
\usepackage{subfigure}
\usepackage{fixme}
\usepackage{amsmath,amsfonts}
\usepackage[ruled]{algorithm2e} 
\usepackage{tensor}

\usepackage{tikz-cd}
\usepackage{tikz}
\usetikzlibrary{positioning}
\usetikzlibrary{arrows}
\tikzset{node distance=1.75cm, auto}

\newcommand{\RR}{\mathbb{R}}

\newcommand{\Id}{\mathrm{Id}}
\newcommand{\argmin}[1]{{\operatorname{argmin}_{#1}}}

\DeclareMathOperator{\Diff}{\mathrm{Diff}}

\newcommand{\ip}[1]{\left\langle #1 \right\rangle}
\newcommand{\pair}[1]{\left\langle #1 \right\rangle}

\usepackage{todonotes}

\begin{document}
\mainmatter              
\title{Bridge Simulation and Metric Estimation on Landmark Manifolds}
\author{Stefan Sommer\inst{1} \and Alexis Arnaudon\inst{2} \and Line Kuhnel\inst{1} \and Sarang Joshi\inst{3}}
%
\authorrunning{Sommer et al.} 

\institute{Department of Computer Science (DIKU), University of Copenhagen, Denmark,
\email{sommer@di.ku.dk}
\and
Department of Mathematics, Imperial College London, UK
\and
Department of Bioengineering, Scientific Computing and Imaging Institute, University of Utah, USA
}

\maketitle              

\begin{abstract}
  We present an inference algorithm and connected Monte Carlo based estimation procedures for metric estimation from landmark configurations distributed according to the transition distribution of a Riemannian Brownian motion arising from the Large Deformation Diffeomorphic Metric Mapping (LDDMM) metric. The distribution possesses properties similar to the regular Euclidean normal distribution but its transition density is governed by a high-dimensional PDE with no closed-form solution in the nonlinear case. We show how the density can be numerically approximated by Monte Carlo sampling of conditioned Brownian bridges, and we use this to estimate parameters of the LDDMM kernel and thus the metric structure by maximum likelihood.

\end{abstract}

\section{Introduction}
Finite dimensional landmark configurations are essential in shape analysis and
computational anatomy, both for marking and following anatomically
important areas in e.g. changing brain anatomies and discretely represented curve 
outlines, and 
in being among the simplest non-linear shape spaces. This
simplicity, in particular the finite dimensionality, makes landmarks useful for
theoretical investigations and for deriving algorithms that can subsequently be
generalized to infinite dimensional shape spaces.

While probability distributions in Euclidean space can often be specified
conveniently from their density function, e.g. the normal distribution with the
density $p_{\mu,\Sigma}(x)\propto e^{-\frac12(x-\mu)^T\Sigma^{-1}(x-\mu)}$, the
non-linear nature of shape spaces often rules out closed form functions.
Indeed, a density defined in coordinates will be dependent on the chosen coordinate chart and
thus not geometrically intrinsic, and normalization factors can be inherently
hard to compute. A different approach defines probability distributions as
transition distributions of stochastic processes. Because stochastic
differential equations (SDEs) can be specified locally from their infinitesimal
variations, it is natural to define them in geometric spaces. Belonging to 
this category, the present paper aligns with a range of 
recent research activities on nonlinear SDEs in shape analysis and geometric
mechanics
\cite{vialard_extension_2013,trouve_shape_2012,marsland_langevin_2016,arnaudon_stochastic_2017,arnaudon_geometric_2017}.
\begin{figure}[!t]
\centering
\includegraphics[width = 0.8\textwidth, trim = 0 70 0 80, clip]{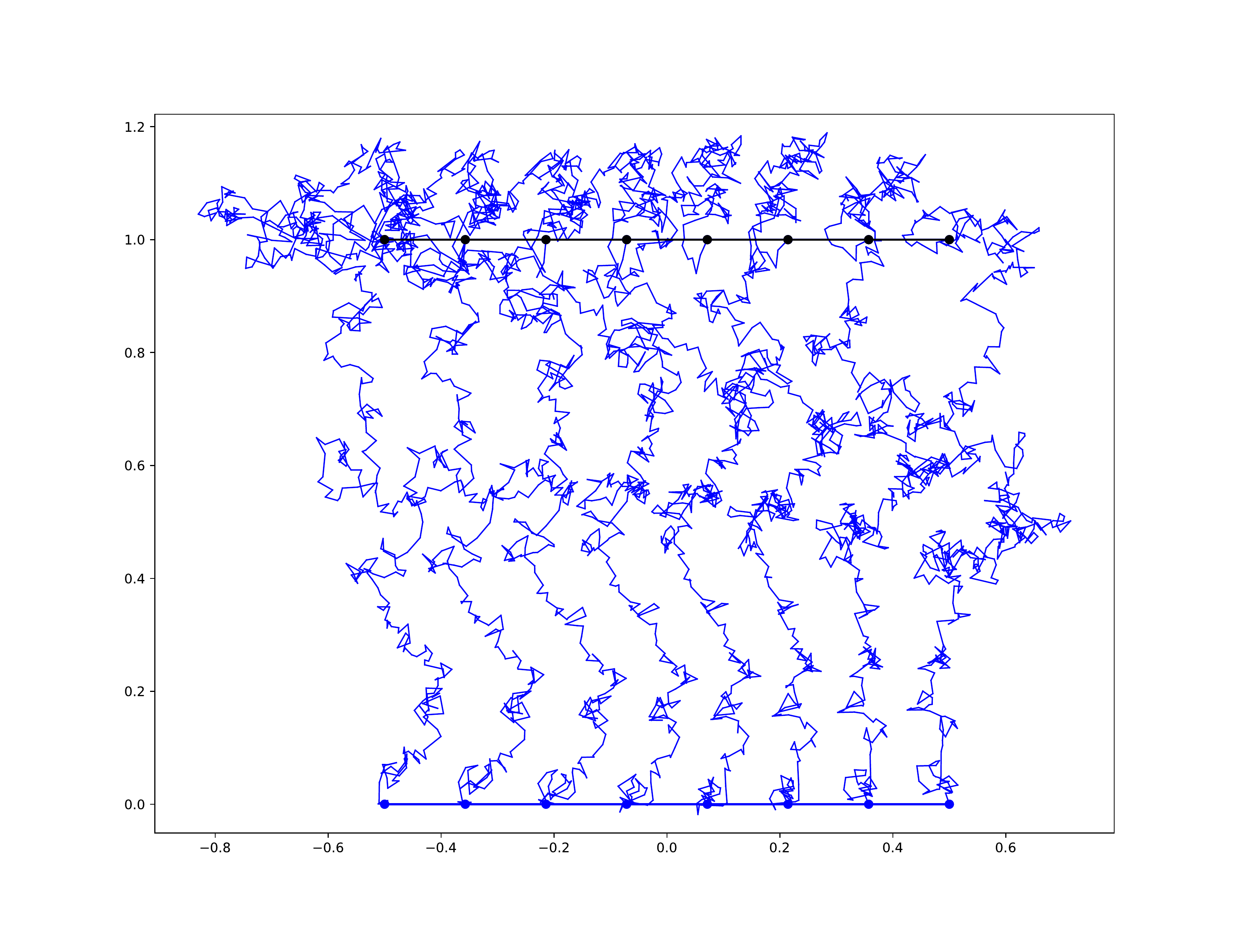}
\caption{A Brownian bridge connecting a configuration of 8 landmarks (blue
points) to corresponding target landmarks (black points). Blue curves shows the
stochastic trajectory of each landmark. The bridge arises
from a Riemannian Brownian motion conditioned on hitting the target at time
$T=1$. The transition density $p_T$ can be evaluated by taking expectation over
such bridges.}
\label{fig:mapping}
\end{figure}

We consider here observations distributed according to the transition
distribution of a Brownian motion, which is arguably one of the most direct
generalizations of the Gaussian distribution to nonlinear geometries.
For the Brownian motion, each infinitesimal step defining the SDE can be
considered normally distributed with isotropic covariance with respect to the
Riemannian metric of the space. Then, from observations, we aim to infer parameters
of this metric. In the Large Deformation Diffeomorphic Metric Mapping
(LDDMM) setting, this can be framed as inferring parameters of the kernel mapping
$K$ between the dual Lie algebra $\mathfrak g^*$ and the Lie algebra 
$\mathfrak g=\mathfrak X(\Omega)$ of the diffeomorphism
group $\Diff(\Omega)$ that acts on the domain $\Omega$ containing the landmarks. 
We achieve this by deriving a scheme for Monte Carlo simulation 
of Brownian landmark
bridges conditioned on hitting the observed landmark configurations. Based on
the Euclidean diffusion bridge simulation scheme of Delyon and Hu
\cite{delyon_simulation_2006}, we can compute expectation over bridges using the correction factor of a guided diffusion process to obtain the transition density of the
Brownian motion. From this, we can take derivatives to obtain an iterative
optimization algorithm for the most likely parameters. The scheme
applies to the situation when the landmark configurations are considered
observed at a fixed positive time $t=T$. 
The time interval $[0,T]$ will generally be sufficiently large that many time
discretization points are needed to accurately represent the stochastic process.

We begin in Section~\ref{sec:background} with a short survey of LDDMM landmark
geometry, metric estimation, large deformation stochastics, and uses of Brownian
motion in shape analysis. In Section~\ref{sec:simulation}, we derive a scheme
for simulating Brownian landmark bridges. We apply this scheme in
Section~\ref{sec:inference} to derive an inference algorithm for estimating
parameters of the metric. Numerical examples are presented in
Section~\ref{sec:experiments} before the paper ends with concluding remarks.

\section{Landmarks Manifolds and Stochastic Landmark Dynamics}
\label{sec:background}
We start with a short survey of landmark geometry with the LDDMM framework as
derived in papers including
\cite{trouve_infinite_1995,dupuis_variational_1998,joshi_landmark_2000,beg_computing_2005}.
The framework applies to general shape spaces though we focus on 
configurations $\mathbf q=(q_1,\ldots,q_N)$ of $N$ landmarks
$q_i\in\Omega\subseteq\RR^d$. We denote the resulting manifold $Q$.
Two sets of shapes $\mathbf q^0,\mathbf q^1$ are in LDDMM matched
by minimizing the energy functional
\begin{equation}
  E(u_t)
  =
  \int_0^1l(u_t)dt
  +
  \frac{1}{2\lambda^2}\|g_1.\mathbf q^0-\mathbf q^1\|^2
  \,.
  \label{eq:E}
\end{equation}
The parameter of $E$ is a time-dependent vector field $u_t\in\mathfrak X(\Omega)$
that via a reconstruction equation
\begin{equation}
  \partial_tg_t
  =
  u_t\circ g_t
  \label{eq:rec}
\end{equation}
generates a corresponding time-dependent flow of diffeomorphisms
$g_t\in\Diff(\Omega)$. The endpoint diffeomorphism $g_1$ move the landmarks
through the action $g.\mathbf q=(g(q_1),\ldots,g(q_N))$ of $\Diff(\Omega)$ on
$Q$. The right-most term of \eqref{eq:E} measures the dissimilarity between 
$g_1.\mathbf q^0$ and $\mathbf q^1$ weighted by a factor $\lambda>0$. In the landmark 
case, the squared Euclidean distance 
when considering the landmarks elements of $\RR^{Nd}$ is often used here.

The Lagrangian $l$ on $u$ is often in the form
$l(u)=\pair{u,Lu}$ with the $L^2$-pairing and $L$ being a differential operator.
Because $\mathfrak X(\Omega)$ can formally be considered the Lie algebra
$\mathfrak g$ of $\Diff(\Omega)$, $l$ puts the dual Lie algebra $\mathfrak g^*$ into
correspondence with $\mathfrak g$ by the mapping 
$\frac{1}{2}\frac{\delta l}{\delta u}:\mathfrak g\rightarrow\mathfrak g^*$, $u\mapsto Lu$. 
The inverse of the
mapping arise from the Green's function of $L$, written as the kernel $K$.
Such $l$ defines a right-invariant inner product on the
tangent bundle $T\Diff(\Omega)$ that descends to a Riemannian metric on $Q$.
Because $Q$ can be considered a subset of $\RR^{Nd}$ using the representation
above, the metric structure can be written directly as a cometric
\begin{equation}
  \ip{\xi,\eta}_\mathbf q
  =
  \xi^T K(\mathbf q,\mathbf q)\eta
  \label{eq:cometric}
\end{equation}
using the kernel $K$ evaluated on $\mathbf q$ for two covectors
$\xi,\eta\in T^*_\mathbf qQ$. The kernel is often specified directly in the form
$K(\mathbf q_1,\mathbf q_2)=\Id_dk(\|\mathbf q_1-\mathbf q_2\|^2)$ for appropriate
kernels $k$. One choice of $k$ is the Gaussian kernel $k(x)=\alpha
e^{-\frac{1}{2}x^T\Sigma^{-1} x}$ with matrix $\Sigma=\sigma\sigma^T$ specifying the spatial 
correlation structure, and $\alpha>0$ a scaling of the general kernel amplitude.
\ \\

Estimating parameters of $K$, with $K$ as above $\alpha$ and the entries of
$\Sigma$ or $\sigma$, has to our knowledge previously only been treated for landmarks in
the small-deformation setting \cite{allassonniere_towards_2007}.
While a linear vector space probability distribution is mapped to the manifold
with small deformations, this paper concerns the
situation when the probability distribution is constructed directly from the
Riemannian metric on the nonlinear space $Q$. The approach has similarities with the
estimation procedures derived in \cite{sommer_brownian_2017} where a metric on a
finite dimensional Lie group is estimated to optimize likelihood of data on a 
homogeneous space arising as the quotient of the group by a closed subgroup. Though 
the landmark space can be
represented as $\Diff(\Omega)/H$ with $H$ the landmark isotropy subgroup
\cite{sommer_reduction_2015}, the approach of \cite{sommer_brownian_2017} can not 
directly be applied because of the infinite dimensionality of $\Diff(\Omega)$.

\subsection{Brownian Motion}
A diffusion processes $\mathbf q_t$ on a Riemannian manifold $Q$ is said to be a 
Brownian motion 
if its generator is $\frac{1}{2}\Delta_g$ with $\Delta_g$ being the Laplace-Beltrami
operator of the metric $g$. Such processes can be constructed in several ways, see
e.g. \cite{hsu_stochastic_2002}. By isometrically embedding $Q$ in a Euclidean
space $\RR^p$, the process can be constructed as a process in $\RR^p$
that will stay on $Q$ a.s. The process can equivalently be characterized 
in coordinates as being solution to the It\^o integral
\begin{equation}
  dq_t^i
  =
  g^{kl}\Gamma(\mathbf q_t)\indices{_{kl}^i}dt
  +
  \sqrt{g^*(\mathbf q_t)}^idW_t
  \label{eq:brown-coords}
\end{equation}
where $\sqrt{g^*}$ is a square root of the cometric tensor $[g^*]^{ij}=g^{ij}$, 
and the drift term arise from contraction of the Christoffel symbols 
$\Gamma\indices{_{kl}^i}$ with the cometric. The noise term is infinitesimal
increments $dW$ of an $\RR^{\mathrm{dim}(Q)}$-valued Brownian motion $W_t$. 
Equivalently, the
Brownian motion can be constructed as a hypoelliptic diffusion processes in the
orthonormal frame bundle $OQ$ where a set of globally defined horizontal
vector fields $H_1,\ldots,H_{\mathrm{dim}(Q)}\in TOQ$ gives the Stratonovich
integral equation
\begin{equation}
  du_t
  =
  H_i(u_t)\circ W_t^i
  \ .
  \label{eq:brown-om}
\end{equation}
Note the sum over the $\mathrm{dim}(Q)$ horizontal fields $H_i$.
The process $\mathbf q_t=\pi(u_t)$ where $\pi:OQ\rightarrow Q$ is the bundle map is then
a Brownian motion. This is known as the Eells-Elworthy-Malliavin
construction of Brownian motion. The fields $H_i$ evaluated at $u\in OQ$ model infinitesimal parallel
transport of the vectors comprising the frame $u$ in the direction of the 
$i$th frame vector, see e.g. \cite{hsu_stochastic_2002}.

While Brownian motion is per definition isotropic with equal variation in all
directions, data with nontrivial
covariance can be modeled by defining the SDE \eqref{eq:brown-om} in the larger
frame bundle $FQ$ 
\cite{sommer_anisotropic_2015,sommer_modelling_2016} using nonorthonormal frames to
model the square root of the local covariance structure. In this setup, the inference 
problem consists of finding the starting point of the diffusion and the
square root covariance matrix.
Estimators are defined via a Frech\'et mean like minimization in $FQ$ with
square $FQ$ distances used as proxy for the negative log-transition density. In this
paper, we remove this proxy by approximating the actual transition density,
but only in the isotropic Brownian motion case.

\subsection{Large Deformation Stochastics}
Several papers have recently derived models for Brownian motion \cite{markussen_large_2007}
and stochastic dynamics in shape analysis and for
landmark manifolds. \cite{trouve_shape_2012,vialard_extension_2013} considered
stochastic shape evolution by adding finite and infinite dimensional noise in
the momentum equation of the dynamics. In \cite{marsland_langevin_2016}, noise
is added to the momentum equation to make the landmark dynamics correspond to a
type of heat bath appearing in statistical physics. In 
\cite{arnaudon_stochastic_2017,arnaudon_geometric_2017} a stochastic model for
shape evolution is derived that descends to the landmark space in the same
fashion as the right-invariant LDDMM metric descends to $Q$. The fundamental
structure is here the momentum map that is preserved by the introduction of
right-invariant noise. The approach is linked to parametric SDEs in fluid
dynamics \cite{holm_variational_2015} and stochastic coadjoint motion
\cite{arnaudon_noise_2016}.

\section{Brownian Bridge Simulation}
\label{sec:simulation}
Brownian motion can be numerically simulated on $Q$ using the coordinate It\^o
form \eqref{eq:brown-coords}. With a standard Euler discretization, the scheme
becomes
\begin{equation}
  \mathbf q_{t_{k+1}}
  =
  \mathbf q_{t_k}
  +
  K(\mathbf q_{t_k},\mathbf q_{t_k})^{kl}\Gamma(\mathbf q_t)\indices{_{kl}}\Delta t
  +
  \sqrt{K(\mathbf q_{t_k},\mathbf q_{t_k})}_j\Delta W_{t_k}^j
  \label{eq:brown-coords-disc}
\end{equation}
with time discretization $t_1,\ldots,t_k$, $t_k-t_{k-1}=\Delta t$ and discrete noise
$W_{t_1},\ldots,W_{t_k}\in\RR^{Nd}$, $\Delta W_{t_k}=\Delta W_{t_k}-\Delta
W_{t_{k-1}}$. Alternatively, a Heun scheme for discrete integration of the
Stratonovich equation \eqref{eq:brown-om} results in
\begin{equation}
  \begin{split}
    &
    v_{t_{k+1}}
    =
    H_i(u_{t_k})\Delta W_{t_k}^i
    \\&
    u_{t_{k+1}}
    =
    u_{t_k}
    +
    \frac{v_{t_{k+1}}+H_i(u_{t_k}+v_{t_{k+1}})\Delta W_{t_k}^i}{2}
  \ .
  \end{split}
\label{eq:brown-om-disc}
\end{equation}
Because the horizontal fields represent infinitesimal parallel transport, they can
be expressed using the Christoffel symbols of $g$. 
The Christoffel symbols for the landmark metric are derived in
\cite{micheli_differential_2008} from which they can be directly implemented or
implicitly retrieved from an automatic symbolic differentiation as done in
the experiments in Section~\ref{sec:experiments}.

\subsection{Bridge Sampling}
The transition density $p_T(\mathbf v)$ 
 of a Brownian motion $\mathbf q_t$ evaluated at 
$\mathbf v\in Q$ at time $T>0$ can be informally obtained by taking an expectation to get the ``mass''
of those of the sample paths hitting $\mathbf v$ at time $T$. We write
$\mathbf q_t|\mathbf v$ for the process $\mathbf q_t$ conditioned on
hitting $\mathbf v$ a.s. at $t=T$.
Computing the expectation analytically is in nonlinear situations generally intractable.
Instead, we wish to employ a Monte Carlo approach and thus derive a method for
simulating from $\mathbf q_t|\mathbf  v$. For this, we employ the bridge sampling scheme of
Delyon and Hu \cite{delyon_simulation_2006}. We first describe the framework for
a general diffusion process in Euclidean space before using it directly on the
landmark manifold $Q$.

Let
\begin{equation}
  dx_t=
  b(t,x_t)dt
  +
  \sigma(t,x_t)dW_t
  \label{eq:driftsde}
\end{equation}
be an $\RR^k$ valued It\^o diffusion with invertible diffusion field
$\sigma$. In order to sample from the conditioned process $x_t|v$, $v\in\RR^k$, a modified
processes is in \cite{delyon_simulation_2006} constructed by adding an extra
drift term to the process giving the new process
\begin{equation}
  dy_t=
  b(t,y_t)dt
  -
  \frac{y_t-v}{T-t}dt
  +
  \sigma(t,y_t)dW_t
  \ .
  \label{eq:guideddriftsde}
\end{equation}
As $t\rightarrow T$, the attraction term $-(y_t-v)/(T-t)dt$ becomes
increasingly strong forcing the processes to hit $y$ at $t=T$ a.s. It can be
shown that the process $y_t$ exists when $b$, $\sigma$ and $\sigma^{-1}$ are
$C^{1,2}$ with bounded derivatives. The process is then absolutely continuous 
with respect to the conditioned process $x_t|v$. The Radon-Nikodym derivative 
between the laws $P_{x_t|v}$ and $P_y$ is
\begin{equation*}
  \frac{dP_{x|v}}{dP_y}(y)
  =
  \frac{\varphi_T(y)}{E_{y}[\varphi_T]}
\end{equation*}
with $E_{y}[\cdot]$ denoting expectation with respect to $P_y$, and the
correction factor $\varphi_T(y)$ defined as the $t\rightarrow T$ limit of
\begin{equation}
\begin{split}
  \varphi_t(y)
  &=
  \exp
  \left(
  -\int_0^t
  \frac{\tilde{y}_s^TA(s,y_s)b(s,y_s)}{T-s}ds
  \right.
  \\
  &\qquad\qquad
  \left.
  -
  \frac{1}{2}
  \int_0^t
  \frac{\tilde{y}_s^T(dA(s,x_s))\tilde{y}_s+\sum_{i,j}d\left<A^{ij}(s,y_s),\tilde{y}_s^i\tilde{y}_s^j\right>}{T-s}
  \right)
\end{split}
  \label{eq:phit}
\end{equation}
Here $\tilde{y}_t=y_t-v$, $A=(\sigma\sigma^T)^{-1}$, and
quadratic variation is denoted by $\left<\cdot,\cdot\right>$. Then
$E_{x}[f(x)|x_T=v]=E_{x|v}[f(x)]=E_{y}[f(y)\varphi_T(y)]/E_{y}[\varphi_T(y)]$
and
\begin{equation*}
  E[f(y)\varphi_t]
  =
  \frac{T^{k/2}e^{\frac{\|\sigma^{-1}(0,x_0)(x_0-v)\|^2}{2T}}}{(T-t)^{k/2}}
  E\big[
    f(x) e^{-\frac{\|\sigma^{-1}(t,x_t)(x_t-v)\|^2}{2(T-t)}}
  \big]
\end{equation*}
for $t<T$.
The fact that the diffusion field $\sigma$ must be invertible for the scheme to
work as outlined here can be seen explicitly from the 
use of the inverse of $\sigma$ and $A$ in these equations.

We can use the guided process \eqref{eq:guideddriftsde} to take conditional expectation for
general measurable functions on the Wiener space of paths $W(\RR^k,[0,T])$ by
sampling from $y_t$. Taking the particular choice of the constant function, the expression
\begin{equation}
  p_T(v)
  =
  \sqrt{\frac{|A(T,v)|}{(2\pi T)^k}}
  e^{\frac{-\|\sigma(0,x_0)^{-1}(x_0-v)\|^2}{2T}}
  E_{y}[\varphi_T(y)]
  \label{eq:eucptapproxgeneral}
\end{equation}
for the transition density of $x_t$ arise as shown in 
\cite{papaspiliopoulos_importance_2012}. Note that both the leading
factors and the correction factor $\varphi_T$ are dependent on the diffusion field
$\sigma$, the starting point $x_0$ of the diffusion, and the drift $b$.
Again, we can approximate the expectation in \eqref{eq:eucptapproxgeneral} by
sampling from $y_t$.

\subsection{Landmark Bridge Simulation}
Because the landmark manifold has a global chart on $\RR^{Nd}$ from the
standard representation of each landmark position in $\RR^d$, we can
conveniently apply the bridge construction of \cite{delyon_simulation_2006}.
Writing the It\^o coordinate form of the Brownian motion $\mathbf q_t$
\eqref{eq:brown-coords} in the form \eqref{eq:driftsde}, we have
$b(t,\mathbf q)=K(\mathbf q,\mathbf q)^{kl}\Gamma(\mathbf q)\indices{_{kl}}$ and 
$\sigma(t,\mathbf q)=\sqrt{K(\mathbf q,\mathbf q)}$ giving the guided SDE
\begin{equation}
  d\mathbf y_t=
  K(\mathbf y_{t_k},\mathbf y_{t_k})^{kl}\Gamma(\mathbf y_t)\indices{_{kl}}dt
  -
  \frac{\mathbf y_t-\mathbf v}{T-t}dt
  +
  \sqrt{K(\mathbf y_{t_k},\mathbf y_{t_k})}dW_t
  \label{eq:guided-landmarks}
\end{equation}
The attraction term $-(\mathbf y_t-\mathbf v)/(T-t)dt$ is the
difference between the current landmark configuration $\mathbf y_t$ and
the target configuration $\mathbf v$. The transition density becomes
\begin{equation}
  p_{T,\theta}(\mathbf v)
  =
  \frac{1}{\sqrt{|K(\mathbf v,\mathbf v)|(2\pi T)^{Nd}}}
  e^{-\frac{\|(\mathbf q_0-\mathbf v)^TK(\mathbf q_0,\mathbf q_0)^{-1}(\mathbf q_0-\mathbf v)\|^2)}{2T}}
  E_{\mathbf y_\theta}[\varphi_{\theta,T}(\mathbf y)]
  \label{eq:trans-dens-landmarks}
\end{equation}
where we use the subscript $\theta$ to emphasize the dependence on the parameters
$\mathbf q_0$ and the kernel $K$. As above, the expectation
$E_{\mathbf y_\theta}[\varphi_{\theta,T}(\mathbf y)]$ can be approximated by
drawing samples from $\mathbf y_\theta$ and evaluating
$\varphi_{\theta,T}(\mathbf y)$.

\begin{remark}
  A similar scheme is used for the bridge simulation of the stochastic coadjoint
  processes of \cite{arnaudon_stochastic_2017,arnaudon_geometric_2017}. In these
  cases, the flow is hypoelliptic in the phase space $(\mathbf q,\mathbf p)$ and 
  observations are partial in that only the landmark positions $\mathbf q$ are
  observed. The momenta $\mathbf p$ are unobserved. In addition, the fact that the landmarks can carry a large initial
  momentum necessitates a more general form of the guidance term
  $-(\mathbf y_t-\mathbf v)/(T-t)dt$ that takes into account the expected value
  of $E_{\mathbf y}[\mathbf y_T|(\mathbf q_t,\mathbf p_t)]$ of the process at
  time $T$ given the current time $t$ position 
  and momentum of the process.
\end{remark}

\section{Inference Algorithm}
\label{sec:inference}
Given a set of i.i.d. observations $\mathbf q^1,\ldots,\mathbf q^N$ of landmark
configurations, we assume
the configurations $\mathbf q^i$ are distributed according to the time $t=T$ transition distribution
$\mathbf q_T$ of a
Brownian motion on $Q$ started at $\mathbf q_0$. We now intend to infer
parameters $\theta$ of the model. With the metric structure on $Q$ given by
\eqref{eq:cometric} and kernel of the form 
$K(\mathbf q_1,\mathbf q_2)=\Id_dk(\|\mathbf q_1-\mathbf q_2\|^2)$,
$k(x)=\alpha e^{-\frac{1}{2}x^T\Sigma x}$, 
parameters are the starting position $\mathbf q_0$, $\alpha$, and 
$\Sigma=\sigma\sigma^T$, i.e. $\theta=(\mathbf q_0,\alpha,\sigma)$.

The likelihood of the model given the data with respect to the Lebesgue measure
on $\RR^{Nd}$ is 
\begin{equation}
  \mathcal{L}_\theta(\mathbf q^1,\ldots,\mathbf q^N)
  =
  \prod_{i=1}^N p_{T,\theta}(\mathbf q^i)
  \label{eq:likelihood}
   .
\end{equation}
Using our ability to approximate \eqref{eq:trans-dens-landmarks} by bridge
sampling, we aim to find a maximum-likelihood estimate (MLE) 
$\hat{\theta}\in\argmin{\theta}\mathcal{L}_\theta(\mathbf q^1,\ldots,\mathbf q^N)$.
We do this by a gradient based optimization on $\theta$, see
Algorithm~\ref{alg:inference}.
Note that the likelihood and thus the MLE of $\theta$ are dependent on the
chosen background measure, in this case coming from the canonical chart on
$\RR^{Nd}$.

\begin{algorithm} \DontPrintSemicolon \SetAlgoLined
  \For{$l=1$ until convergence}{ 
    \For{$i=1$ to $N$}{ 
      sample $J$ paths from guided process $\mathbf y_{\theta_l}$ hitting $\mathbf q^i$\\
      compute correction factors $\varphi_{\theta_l,T}^{i,j}$
    }
    $\mathcal{L}_{\theta_l}(\mathbf q^1,\ldots,\mathbf q^N)
    \leftarrow
    \prod_{i=1}^N
\frac{1}{\sqrt{|K(\mathbf q^i,\mathbf q^i)|(2\pi T)^{Nd}}}
  e^{-\frac{\|(\mathbf q_0-\mathbf q^i)^TK(\mathbf q_0,\mathbf q_0)^{-1}(\mathbf
q_0-\mathbf q^i)\|^2)}{2T}}
\frac{1}{J}\sum_{j=1}^J\varphi_{\theta_l,T}^{i,j}$
    \\
    $\theta_{l+1}=\theta_l+\epsilon\nabla_{\theta_l}\mathcal{L}_{\theta_l}(\mathbf
    q^1,\ldots,\mathbf q^N)$
  }
  \caption{Metric estimation: Inference of parameters $\theta$ from samples.}
  \label{alg:inference}
\end{algorithm}

\begin{remark}
  The inference Algorithm~\ref{alg:inference} optimizes the likelihood 
  $\mathcal{L}_\theta$ directly by stochastic gradient
  descent. A different but related approach is an
  Expectation-Maximization approach where the landmark trajectories between $t=0$ and
  the observation time $t=T$ are considered missing data. The $E$-step of the EM algorithm would then
  involve the expectation $E_{\mathbf x|\mathbf q^i}[\log p(\mathbf x)]$ of the 
  landmark bridges conditioned on the data with $p(\mathbf x)$ formally denoting a likelihood
  of an unconditioned sample path $\mathbf x$. This approach is used in e.g.
  \cite{arnaudon_geometric_2017}. While natural to formulate, the approach
  involves the likelihood $p(\mathbf x)$ of a stochastic path which is only
  defined for finite time discretizations. In addition, the expected
  correction factor $E_{\mathbf y}[\varphi_T(\mathbf y)]$ that arise when
  using the guided process $\mathbf y$ in the estimation appears as a
  normalization factor in the EM $Q$-function. This can potentially make the scheme
  sensitive to the
  stochasticity in the Monte Carlo sampling of the expected correction $E_{\mathbf y}[\varphi_T(\mathbf y)]$. While the differences between these
  approaches needs further investigation, we hypothesize that direct 
  optimization of the likelihood is superior in the present context.
\end{remark}

\begin{remark}
  Instead of taking expectations over $\mathbf q_T$, we can identify the most 
  probable path of the conditioned process $\mathbf q_t|\mathbf v$. This results in the
  Onsager-Machlup functional \cite{fujita_onsager-machlup_1982}.
  In \cite{sommer_anisotropically_2016}, a different definition 
  is given that, in the isotropic Brownian motion situation, makes the set of Riemannian geodesics from
  $\mathbf q_0$ to $\mathbf v$ equal to the set of most
  probable paths of the conditioned process $\mathbf q_t|\mathbf v$. The
  sample Frech\'et mean
  \begin{equation}
    \argmin{\mathbf q_0}
    \frac{1}{N}\sum_{i=1}^Nd_g(\mathbf q_0,\mathbf q_i)^2
    \label{eq:frechet-mean}
  \end{equation}
  is in that case formally also a minimizer of the negative log-probability of the most probable
  path to the data. Given that we are now able to approximate the density
  function of the Brownian motion, the MLE of the likelihood
  \eqref{eq:likelihood} with respect to $\mathbf q_0$ is equivalent to
  \begin{equation}
    \argmin{\mathbf q_0}
    -\frac{2}{N}\sum_{i=1}^N\log p_{T,\mathbf q_0}(\mathbf q_i)
    \ .
    \label{eq:density-frechet-mean}
  \end{equation}
  Compared to \eqref{eq:frechet-mean},
 the negative log-probability of the data is here minimized instead of the
  squared geodesic distance. The estimator \eqref{eq:density-frechet-mean} can therefore be
  considered a transition density equivalent of the sample Frech\'et mean.
\end{remark}
\begin{figure}[!t]
\centering
\includegraphics[width = 0.32\columnwidth, trim = 50 70 50 70, clip]{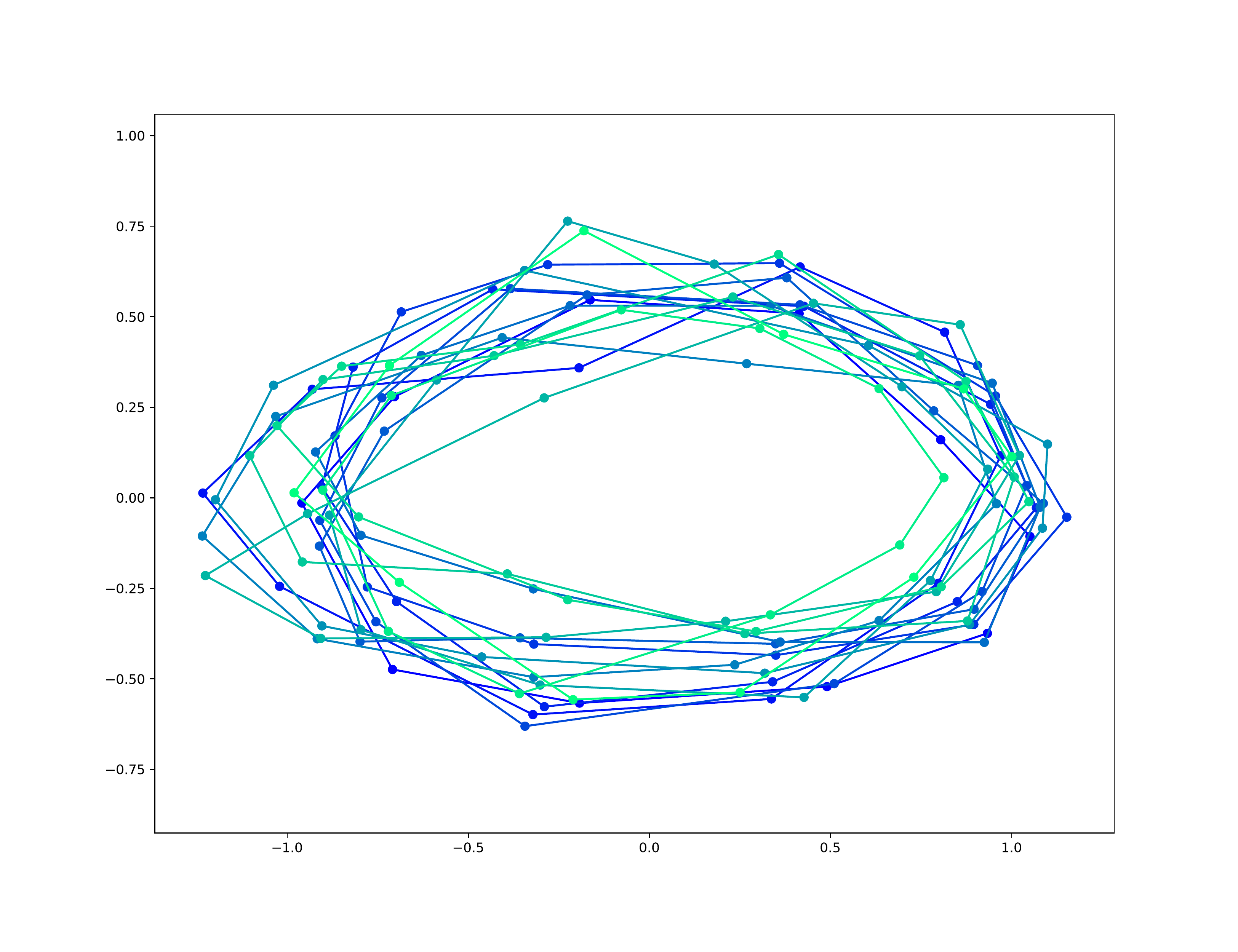}
\includegraphics[width = 0.32\columnwidth, trim = 50 70 50 70, clip]{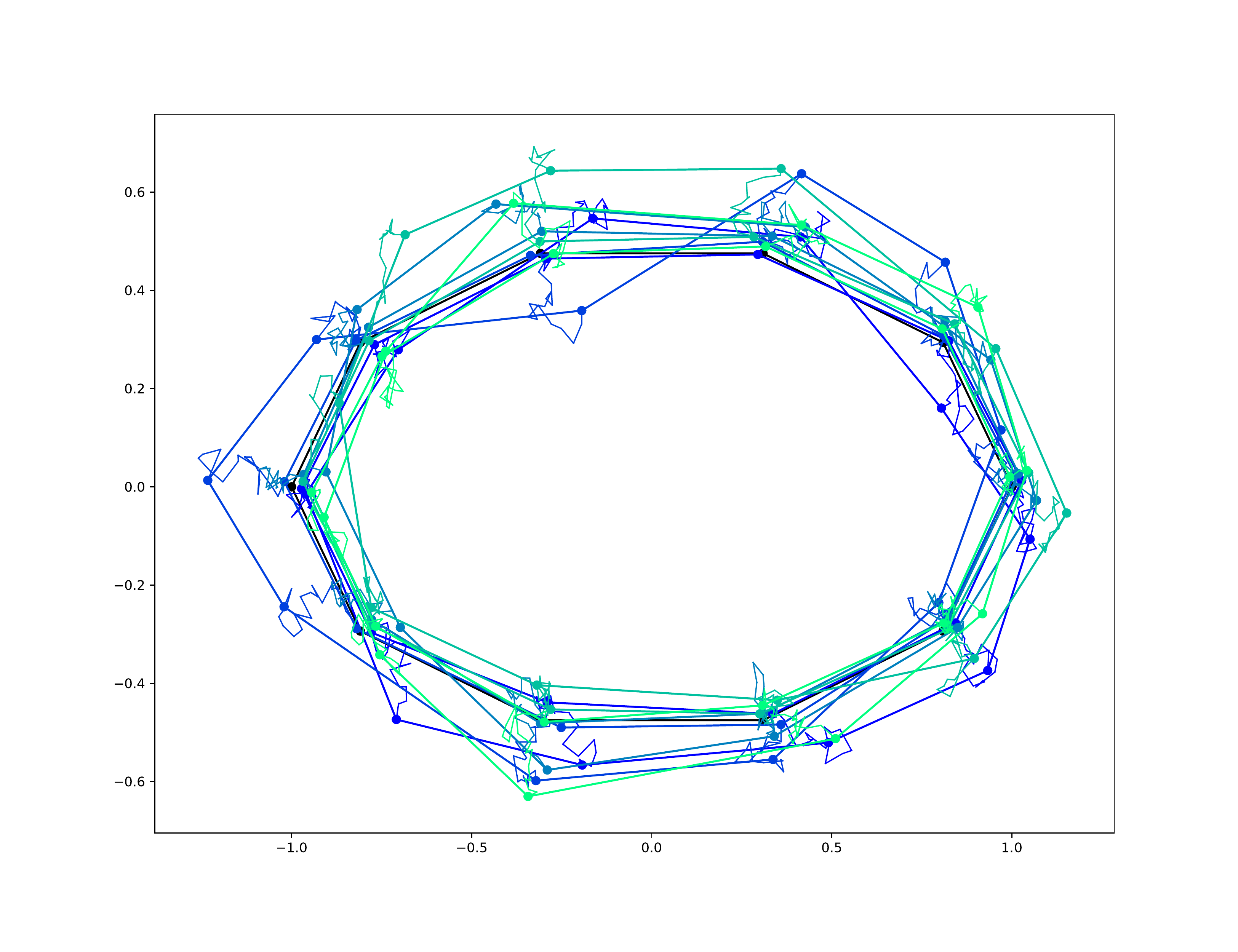}
\includegraphics[width = 0.32\columnwidth, trim = 50 70 50 70, clip]{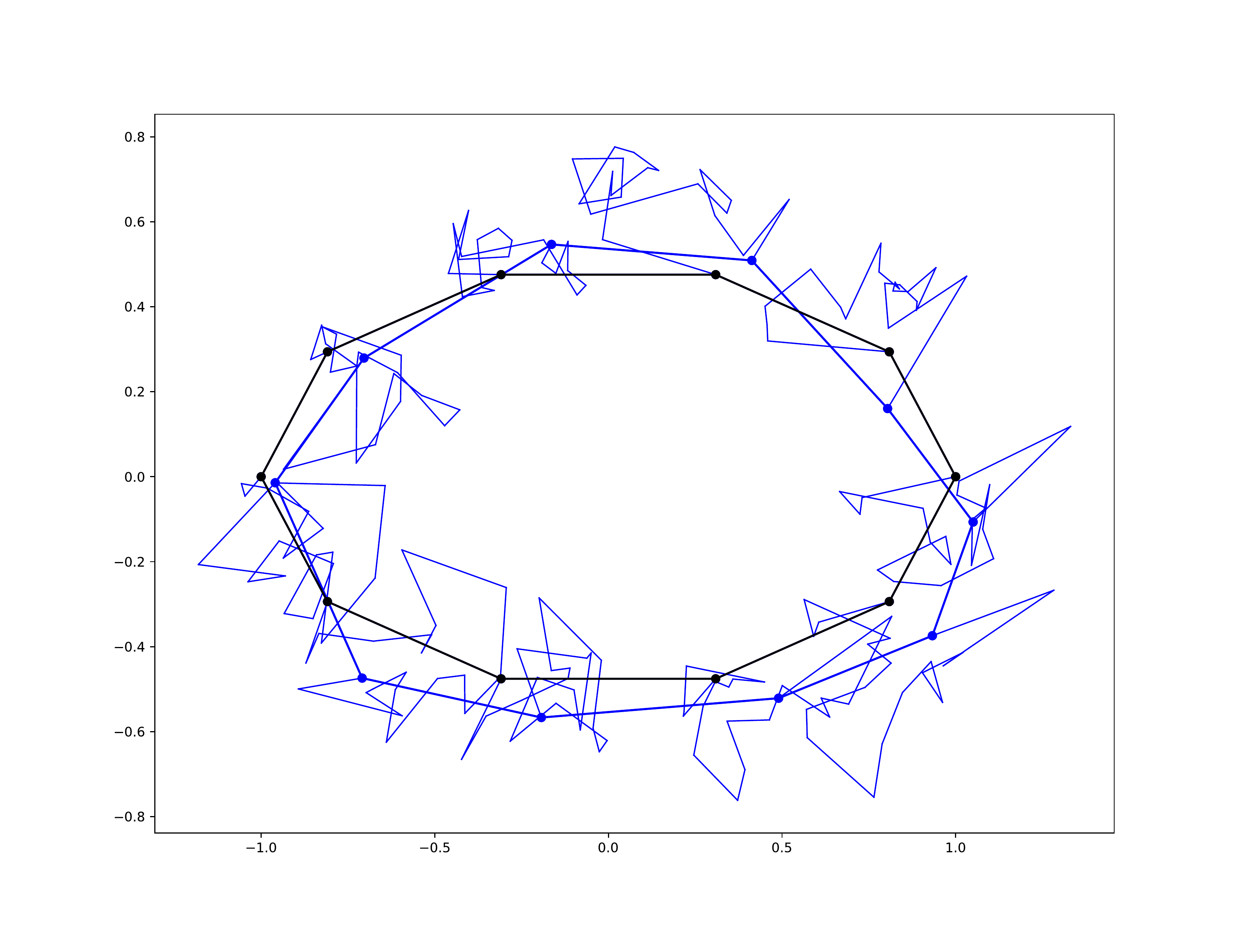}
\caption{(left) Samples from the transition distribution of a Brownian motion
  perturbing an ellipse configuration with 10 landmarks. (center) Subset of the
samples shown together with trajectories of the stochastic landmark Brownian
motion started at configuration $\mathbf q_0$ (black points). (right) A guided
bridge from $\mathbf q_0$ (black points) to a sample (blue points).}
\label{fig:ellipse_samples}
\end{figure}
\begin{figure}[!t]
\centering
\includegraphics[width = 0.36\columnwidth, trim = 80 120 50 100, clip]{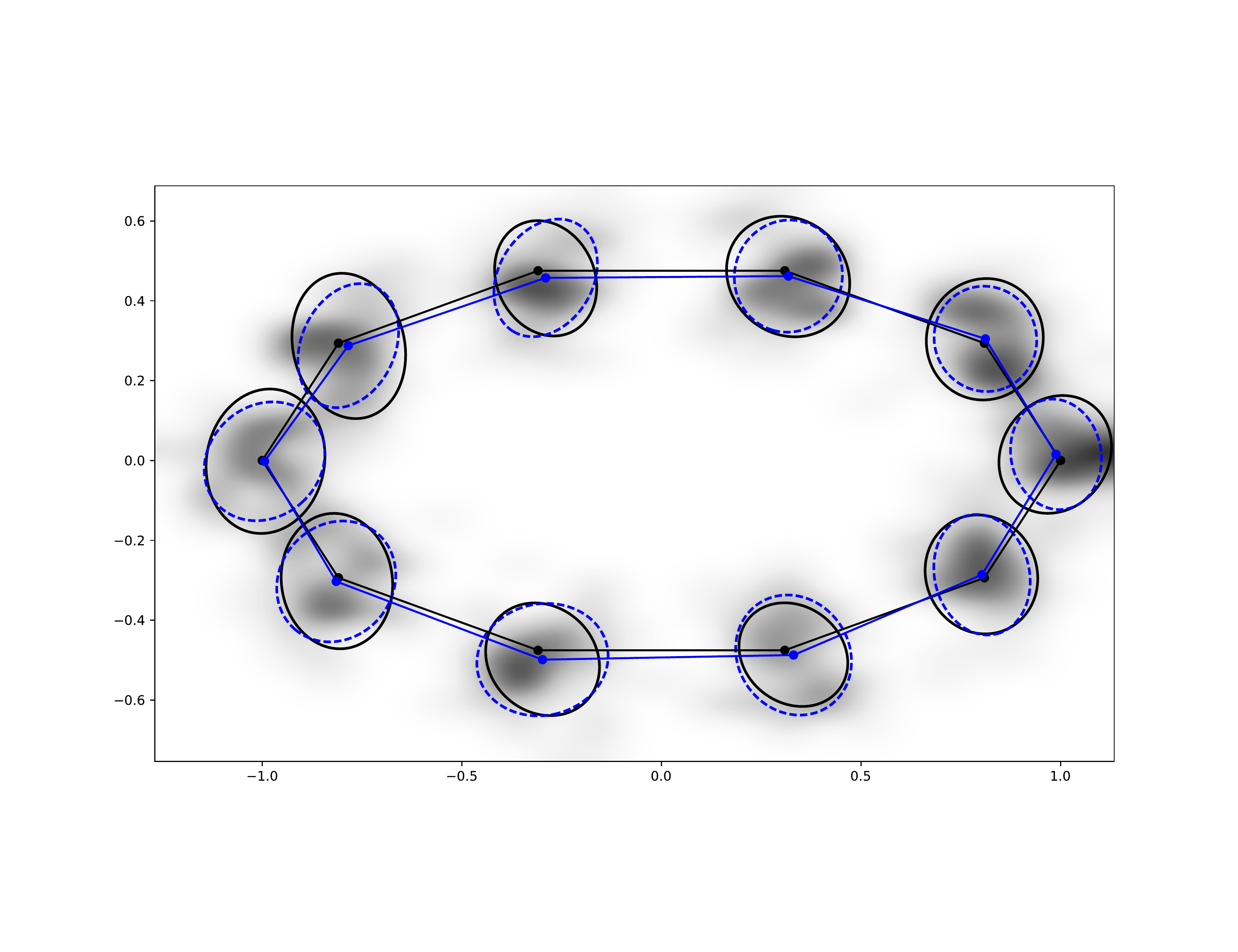}
\includegraphics[width = 0.31\columnwidth, trim = 50 70 35 80, clip]{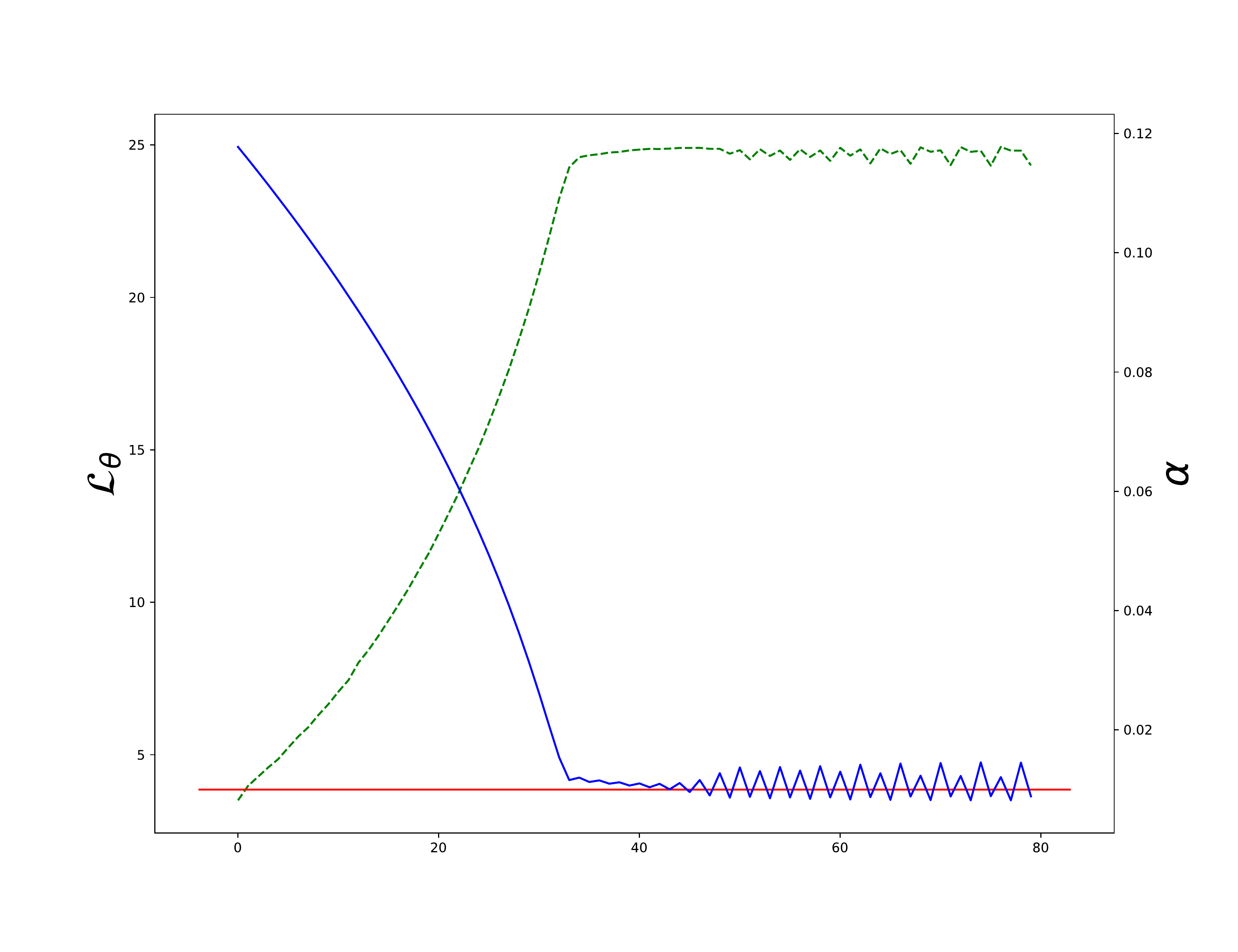}
\includegraphics[width = 0.305\columnwidth, trim = 50 70 50 80, clip]{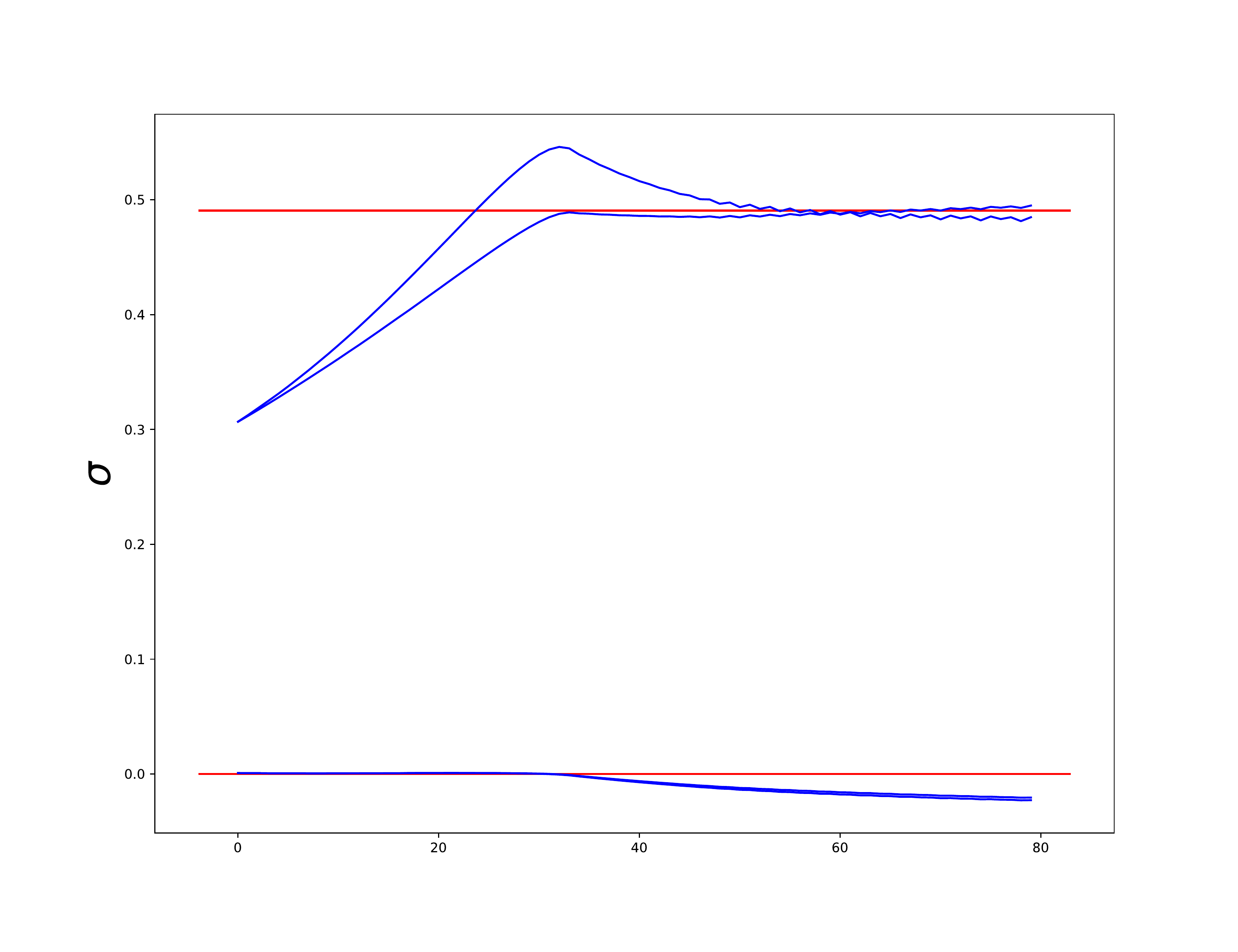}
\caption{(left) Result of the inference algorithm applied to the synthetic
ellipse samples. The initial configuration $\mathbf q_0$ (black points) is
showed along with the per-landmark sample covariance from the samples (black
ellipses). The
estimated initial configuration $\hat{\mathbf q}_0$ (blue points) is shown along
with the per-landmark sample covariance from a new set of samples obtained with
the inferred parameters (blue ellipses). (center) Evolution of likelihood
(green) and 
$\alpha$ (blue) during the
optimization. Horizontal axis shows number of iterations and red line is
$\alpha$ ground
truth. (right) Evolution of the entries of the kernel covariance 
$\sigma$ (blue lines) during the optimization, red lines ground truth.}
\label{fig:ellipse_results}
  \vspace{-.3cm}
\end{figure}
\vspace{-.3cm}

\section{Numerical Experiments}
\label{sec:experiments}
We here present examples of the method on simulated landmark configurations, and
an application of the method and algorithm to landmarks annotated on cardiac
images of left ventricles. We aim for testing
the ability of the algorithm to infer the parameters of the model given samples.
We here take the first steps in this direction and leave a more extensive simulation
study to future work.

For the simulated data, we compare the results against the true values used in the
simulation. In addition, we do simple model checking for both experiments by simulating
with the estimated parameters and comparing the per-landmark sample mean and
covariance. 

We use code based on the Theano library 
\cite{the_theano_development_team_theano:_2016} for the implementation, in particular the
symbolic expression and automatic derivative facilities of Theano. The code used for
the experiments is available in the software package
\emph{Theano Geometry} \url{http://bitbucket.org/stefansommer/theanogeometry}.
The implementation and the use of Theano for differential geometry applications 
including landmark manifolds is described in \cite{kuhnel_deep_2017}.
\begin{figure}[!t]
  \vspace{-.3cm}
\centering
\includegraphics[width = 0.395\columnwidth, trim = 80 120 50 100, clip]{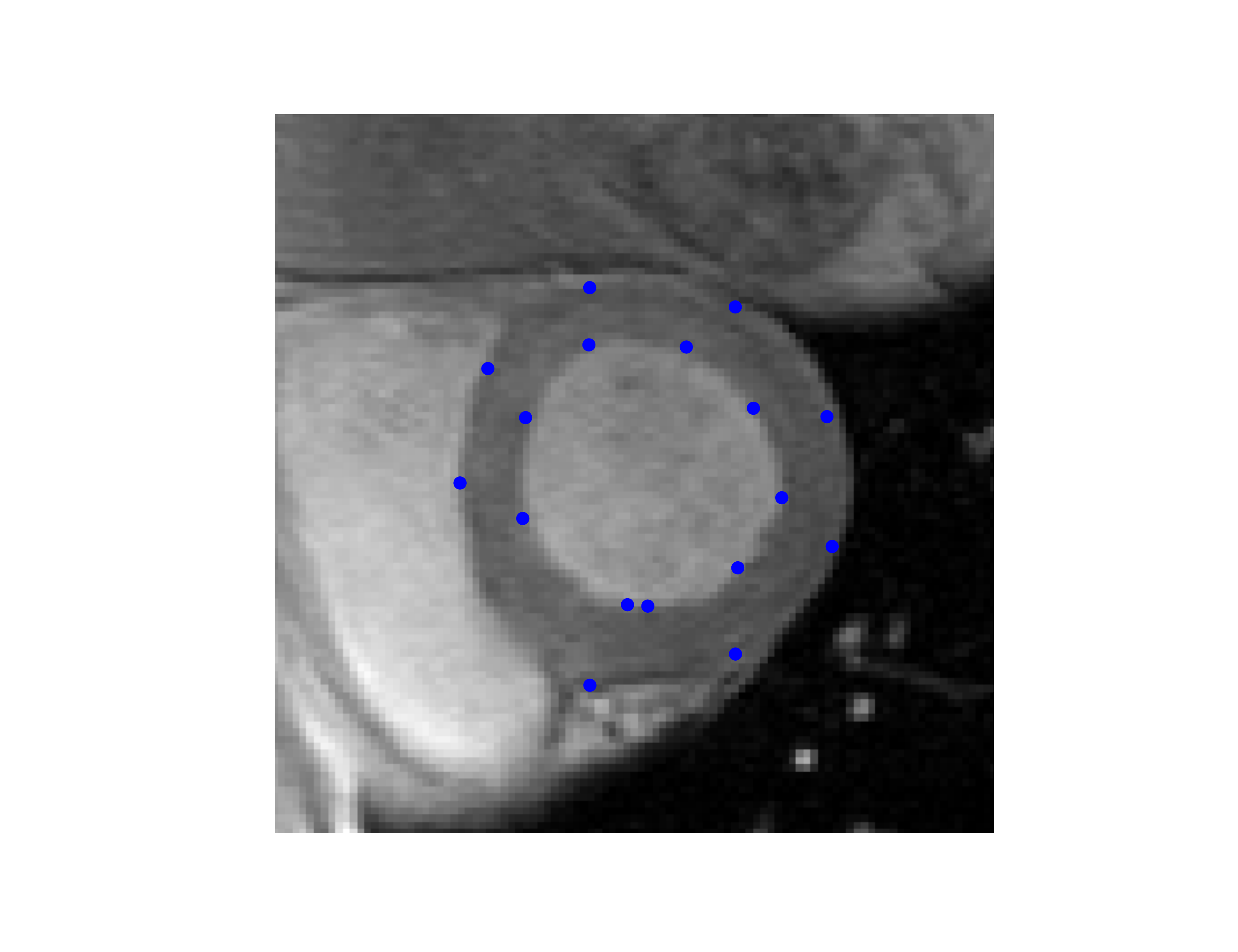}
\includegraphics[width = 0.36\columnwidth, trim = 50 70 50 80, clip]{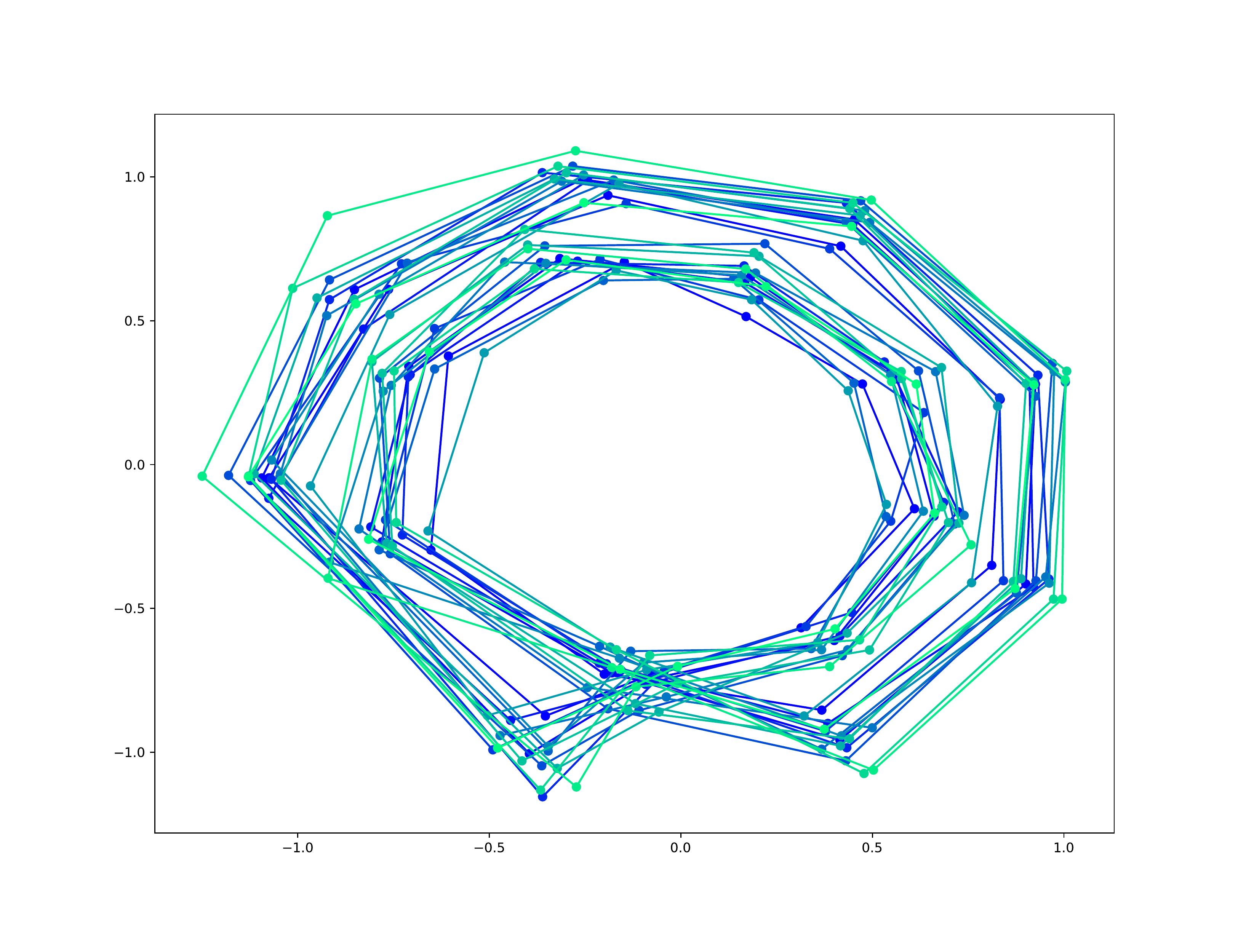}
\caption{(left) 
  An image of a left cardiac ventricle with annotations. (right) The annotations
  from 14 cardiac images.
}
\label{fig:cardiac}
\end{figure}
\begin{figure}[!t]
\centering
\includegraphics[width = 0.36\columnwidth, trim = 80 70 80 80, clip]{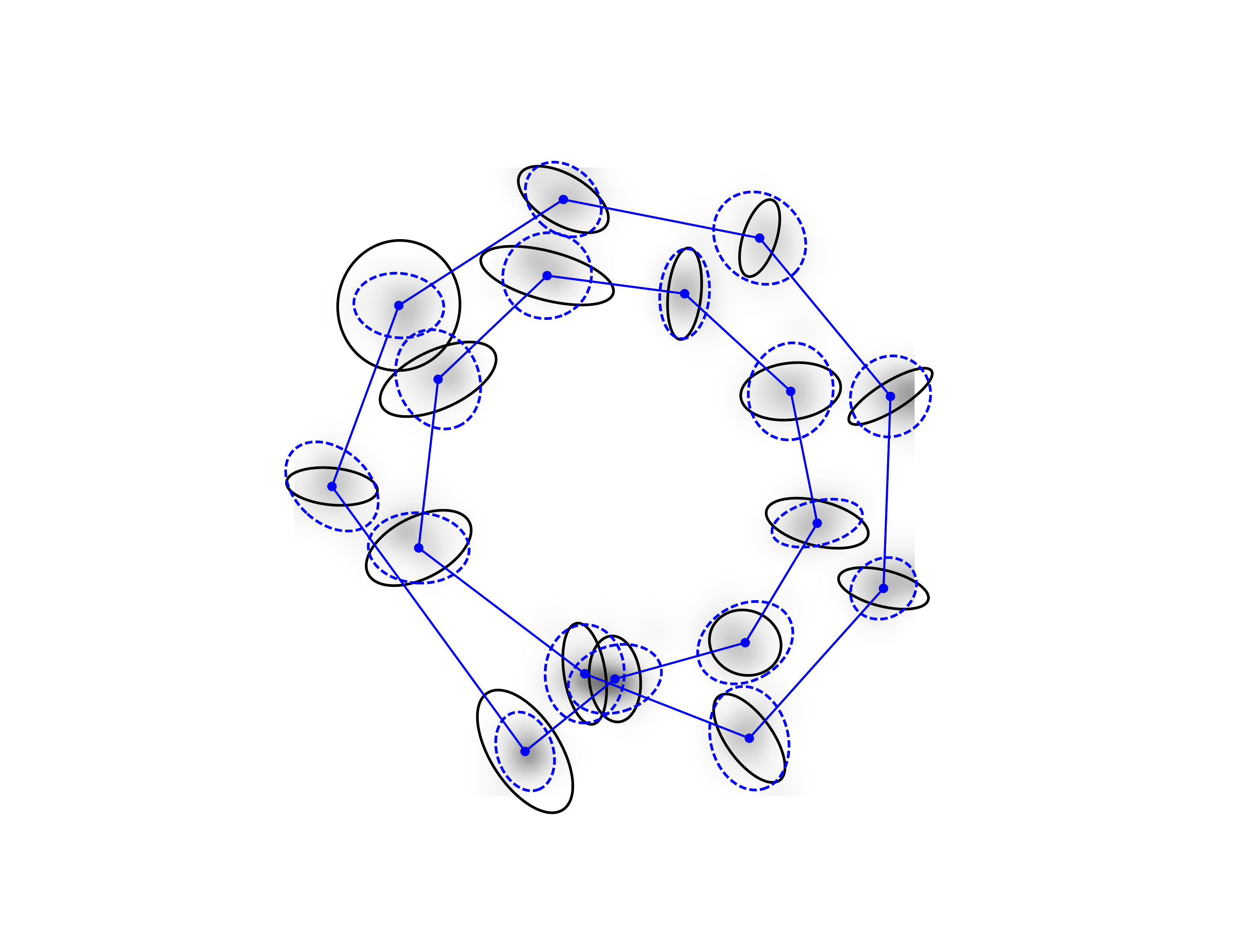}
\includegraphics[width = 0.31\columnwidth, trim = 50 70 50 80, clip]{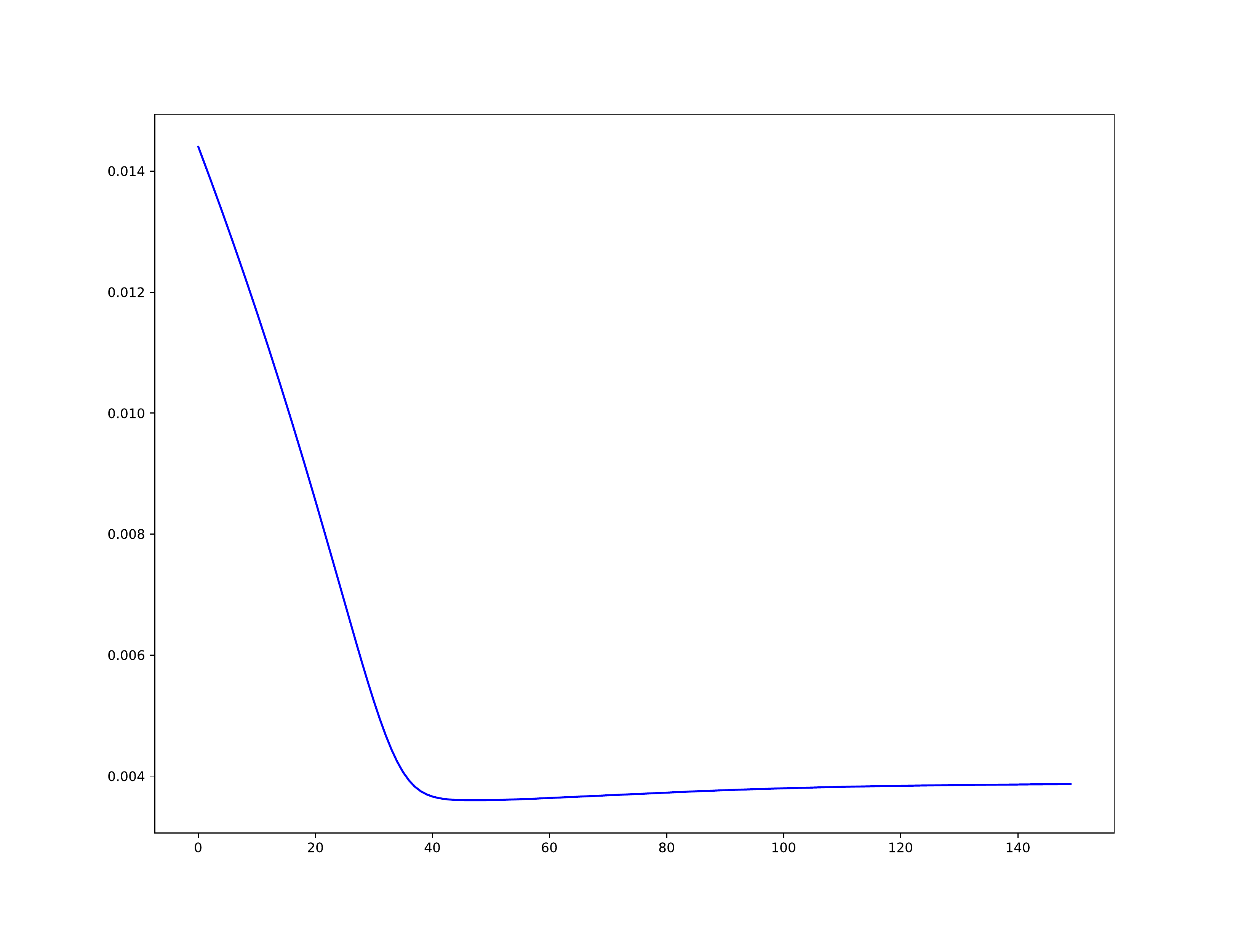}
\includegraphics[width = 0.31\columnwidth, trim = 50 70 50 80, clip]{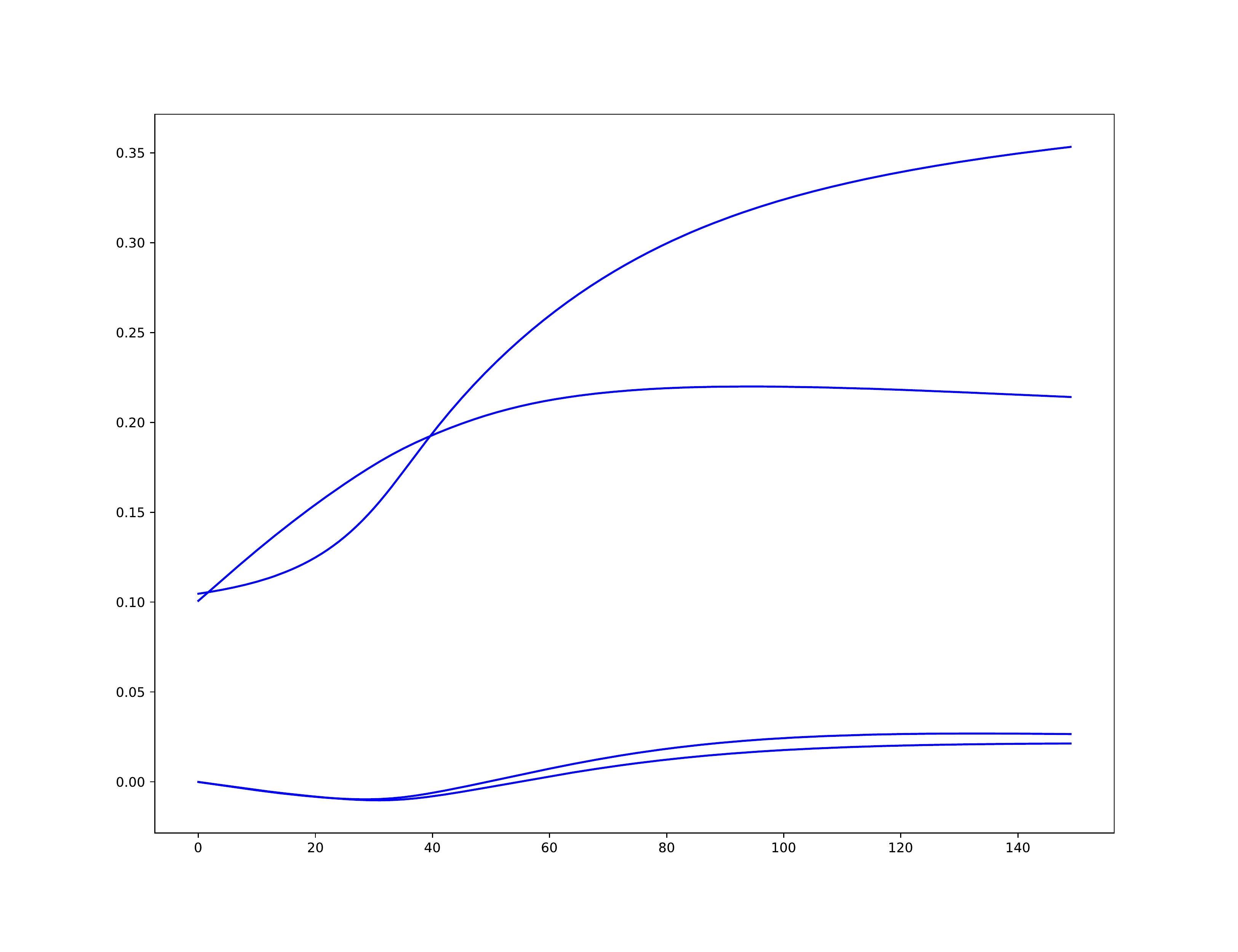}
\caption{Results of applying the inference algorithm to the cardiac data. Setup
  and subfigures as in Figure~\ref{fig:ellipse_results}.}
\label{fig:cardiac_results}
\end{figure}
\ \\

With 10 landmarks arranged in an ellipse configuration $\mathbf q_0$, we sample
64 samples from the transition distribution at time $T=1$ of a Brownian motion
started at $\mathbf q_0$, see Figure~\ref{fig:ellipse_samples}. 
Parameters for the kernel
are $\sigma=\mathrm{diag}(\sigma_1,\sigma_2)$ with $\sigma_1,\sigma_2$ 
set to the average inter-point
distance in $\mathbf q_0$, and the amplitude parameter $\alpha=0.01$. 

We run Algorithm~\ref{alg:inference} with initial conditions for $\mathbf q_0$ the
pointwise mean of the samples. 
The parameter evolution trough the iterative optimization and the result of the
inference can be seen in Figure~\ref{fig:ellipse_results}. The algorithm is able
estimate the initial configuration and the parameters of $\alpha$ and $\Sigma$
with a reasonable precision. The sample per-landmark covariance as measured on a
new set of simulated data with the estimated parameters is comparable to the 
per-landmark covariance of the original dataset.

\subsection{Left Cardiac Ventricles}
To exemplify the approach on real data, we here use a set of landmarks obtained 
from annotations of the left ventricle in 14 cardiac images
\cite{stegmann_extending_2001}. Each ventricle is
annotated with sets of landmarks from which we select 17 from each configuration
for use in this experiment. Figure~\ref{fig:cardiac} shows an annotated image along 
with the sets of annotations for all images.

Figure~\ref{fig:cardiac_results} shows the results of the inference algorithm
with setup equivalent to Figure~\ref{fig:ellipse_results}. While the parameters
converges during the iterative optimization, we here have no ground-truth
comparison. A subsequent sampling using the estimated parameters allows
comparison of the per-landmark sample covariance. While the new sample
covariance in magnitude and to some degree shape corresponds to the sample
covariance from the original data, the fact that the Brownian motion is
isotropic forces the covariance to be equivalent for all landmarks as measured
by the Riemannian landmark metric. Including anisotropic covariance in the
distribution or the right-invariant stochastics of \cite{arnaudon_stochastic_2017,arnaudon_geometric_2017} would allow the per-landmark covariance to vary and result in a
closer fit.

\section{Conclusion}
\label{sec:conclusion}
In the paper, we have derived a method for maximum likelihood estimation of
parameters for the starting point of landmark Brownian motions and for the Riemannian
metric structure specified from the kernel $K$. Using the guided process
scheme of \cite{delyon_simulation_2006} for sampling conditioned Brownian
bridges, the transition density is approximated by Monte Carlo sampling. With
this approximation of the data likelihood, we use a gradient based iterative
scheme to optimize parameters. We show on synthetic and real data sets
the ability of the method to infer the underlying parameters of the
data distribution and hence the metric structure of the landmark manifold.

A direct extension of the method presented here is to generalize to
the anisotropic normal distributions \cite{sommer_anisotropic_2015} defined 
via Brownian motions in the frame
bundle $FQ$. This would allow differences in the per-landmark covariance and
thus improve results on datasets such as the presented left ventricle annotations.
Due to the hypoellipticity of the anisotropic flows that must be
conditioned on hitting fibers in $FQ$ above points $q\in Q$, further work is
necessary to adapt the scheme presented here to the anisotropic case.

\ \\
{\bf Acknowledgements}\quad
We are grateful for the use of the cardiac ventricle dataset provided by Jens Chr.
Nilsson and Bj\o rn A. Gr\o nning, Danish Research Centre for Magnetic Resonance
(DRCMR).

\bibliographystyle{splncs03}
\bibliography{ss}

\end{document}